\documentclass[]{spie}

\usepackage{amsmath,amsfonts,amssymb}
\usepackage{graphicx}
\usepackage[colorlinks=true, allcolors=blue]{hyperref}

\title{Compression Fractures Detection on CT
}

\author[1,2]{Amir Bar}
\author[1]{Lior Wolf}
\author[2]{Orna Bergman Amitai}
\author[2]{Eyal Toledano}
\author[2]{Eldad Elnekave}
\affil[1]{The Blavatnik School of Computer Science, Tel Aviv University} 
\affil[2]{Zebra Medical Vision}

\begin{document} 
\maketitle

\begin{abstract}
The presence of a vertebral compression fracture is highly indicative of osteoporosis and represents the single most robust predictor for development of a second osteoporotic fracture in the spine or elsewhere. Less than one third of vertebral compression fractures are diagnosed clinically. We present an automated method for detecting spine compression fractures in Computed Tomography (CT) scans. The algorithm is composed of three processes. First, the spinal column is segmented and sagittal patches are extracted.  The patches are then binary classified using a Convolutional Neural Network (CNN). Finally a Recurrent Neural Network (RNN) is utilized to predict whether a vertebral fracture is present in the series of patches.
\end{abstract}


\section{INTRODUCTION}
\label{sec:intro}
\subsection{Vertebral Fracture}

The annual incidence of osteoporotic fractures in the United States surpasses the cumulative incidence of breast cancer, heart attack and stroke. The most common osteoporotic fractures are those which result in compression of a vertebral body [\citenum{kanis2004family}]. In addition to being a direct source of morbidity, vertebral compression fractures (VCF’s) indicate substantially elevated risk of future osteoporotic hip fractures, which represent the most consequential sequelae of osteoporosis.  

Within a year of suffering osteoporotic hip fracture, 30\% of previously independent indiviuals will lose the ability to walk without assistance [\citenum{johnell2006estimate}], 25\% will become totally dependent or require nursing home facilities [\citenum{cite3},\citenum{leibson2002mortality}] and 20\% will not survive [\citenum{schnell20101}]. 

Prophylactic treatment of at-risk osteoporotic individuals has been shown to reduce the rate of future hip fracture by 40-70\%[\citenum{kanis2008european},\citenum{black2007once},\citenum{national2008alendronate}]. Yet, despite the burden and potential to prevent hip fractures in an ageing population, osteoporosis screening remains profoundly underutilized: less than 20\% of people older than 65 years undergo bone mineral density screening via Dual Energy X-ray Absorptiometry (DXA) evaluation, [\citenum{cite9}] as recommended by the National Osteoporosis Foundation. 

VCF detection on CT examination is as predictive for future osteoporotic hip fractures as a DXA diagnosis of osteoporosis: detection of a VCF confers a relative risk ratio of 2.3 for experiencing a hip fracture and a 4.4 fold risk of incurring other additional insufficiency fractures [\citenum{national2008alendronate},\citenum{siris2001identification},\citenum{lindsay2001risk},\citenum{roux2007mild}]. Yet only 13\%-16\% of retrospectively confirmed VCF’s are actually reported at the time of computed tomographic (CT) interpretation [\citenum{carberry2013unreported}, \citenum{bartalena2009prevalence},\citenum{williams2009under}]. The expertise required to diagnose VCF's is far less than that required to make the vast majority of routine radiologic observations on CT imaging. In Carberry et al.’s study of VCF's, examining over 2000 CT examinations, the retrospective detection was accomplished by a medical student with one hour of dedicated compression fracture detection training [\citenum{carberry2013unreported}].  The reason why VCF's are routinely missed is more likely due to the fact that they are considered incidental findings relative to the primary clinical indication which prompted the CT study. 

Here we describe a method to increase identification of at-risk individuals via automatic opportunistic detection of VCF’s on CT imaging. 

\subsection{Convolutional Neural Networks}
Convolutional Neural Networks (CNN's) have proven immense utility when applied to Computer Vision tasks such as detection, segmentation and classification. Among CNN's advantages is the ability to learn a hierarchical representation from the input images and extract relevant features that generalize well across a large volume of data. This stands in contrast to conventional methods reliant on handcrafted feature design. Among the most successful CNN architectures for 2D image inputs are AlexNet, VGG, ResNet and Inception [\citenum{simonyan2014very},\citenum{krizhevsky2012imagenet}, \citenum{DBLP:journals/corr/HeZRS15}, \citenum{DBLP:journals/corr/SzegedyVISW15}]. 

\subsection{Recurrent Neural Networks}
Recurrent Neural Networks (RNN's) are a key Deep Learning tool used to model sequences and time series. RNN's are often combined with CNN's, using the CNN as a feature extractor and the RNN to model the sequence.  RNN's are commonly utilized in tasks such as video classification [\citenum{DBLP:journals/corr/NgHVVMT15}], language modeling [\citenum{mikolov2010recurrent}], and speech recognition [\citenum{graves2013speech}], among others. 

The algorithm proposed here comprises three main steps, once provided a CT of the chest and/or abdomen. First, a segmentation process extracts sagittal patches along the vertebral column. These patches are then binary classified using a CNN. Finally, an RNN is run on the resulting vector of probabilities. The RNN output is a prediction of the probability for the presence of a compression fracture within the input CT scan.

\section{RELATED WORK}

Previous work on spine segmentation and VCF detection focused primarily on achieving an accurate  segmentation of each vertebra. Yao et al [\citenum{yao2012detection}] defined a novel method for segmentation of vertebrae, extracting the vertebral cortical circumference and mapping it into 2D for fracture detection.  Ghosh et al [\citenum{ghosh2011automatic}] describe a VCF detection method based upon a 2D sagittal reconstruction, with relatively unstable results.  Kelm et al [\citenum{kelm2013spine}] describe vertebral segmentation method using a disc-centered approach based on a probabilistic model which requires the full vertebral column to be included in the scan. Yao et al [\citenum{yao2006automated}] accomplishes the segmentation on axial slices – assessing the vertebral body as compared to an axial vertebral model. Discs are detected by low similarity to the model. This method was later used for feature-extraction based classification of vertebral fractures of osteoporotic or neoplastic etiology [\citenum{DBLP:journals/corr/WangYBS16}].

\section{DATA} \label{data}

We assembled an initial dataset from 3701 individuals over the age of 50 years who had undergone CT scans of Chest and/or Abdomen for any clinical indication. Two expert radiologists reviewed each CT and assessed them for the presence of VCF's as defined by Genant criteria for vertebral compression. A third radiologist (EE) served to mediate in instances of non-consensus. Of the 3701 CT studies,  2681 (72\%)  were designated as negative for the presence of VCF and 1020 (28\%) were tagged as VCF positive, including a bounding box annotation indicating the fracture position.

The positive class was comprised of 61\% women of an average age of 73 years (std. 12.4), and 39\% men of average age 66.8 years (std. 16.8). The negative class was comprised of 47\% women of average age 56.7 years (std. 17.4) and 53\% men of average age 56.1 years (std. 17.9). These results are not surprising: the prevalence of VCF's is known to be substantially higher in women and elderly individuals. We determined however that such pronounced inter-group differences in demographic characteristics could introduce biases with unintended results.  Training a classifier on this dataset might, for example, have resulted in an algorithm which has learned to distinguish between the spines of elderly women and younger men. 

Considering the above, a more demographically balanced subset of CT studies was derived including 1673 CT studies, of which 849 were VCF negative and 824 VCF positive. The VCF negative and VCF positive groups were filtered to balance age and sex.  The negative class contains 43\% men of average age of 64.7 years (std. 15.9) and 57\% women of average age years 69.4 (std. 11.8). The positive class contains 42\% men of average age 65.7 years (std. 16.0) and 58\% women of average age 70.4 years (std. 11.9). All algorithmic training was performed upon this more balanced data set. 

\begin{figure} [t!]
   \begin{center}
   \begin{tabular}{c}
   \includegraphics[height=4.5cm]{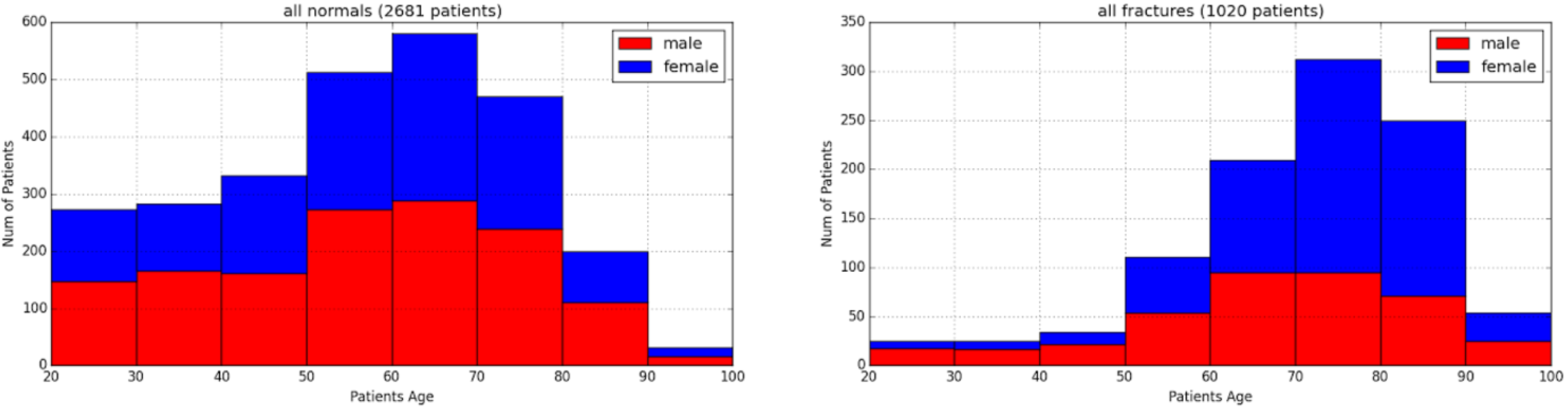}
   \end{tabular}
   \end{center}
   \caption[dataset] 
   { \label{fig:init_dataset} 

Initial dataset: negative class (left) and positive class (right). The positive class average age and the portion of females is higher. Training on this might result in learning a classifier which heavily relies on age and gender rather than VCF's. }
   \end{figure}

\begin{figure} [t!]
   \begin{center}
   \begin{tabular}{c}
   \includegraphics[height=4.6cm]{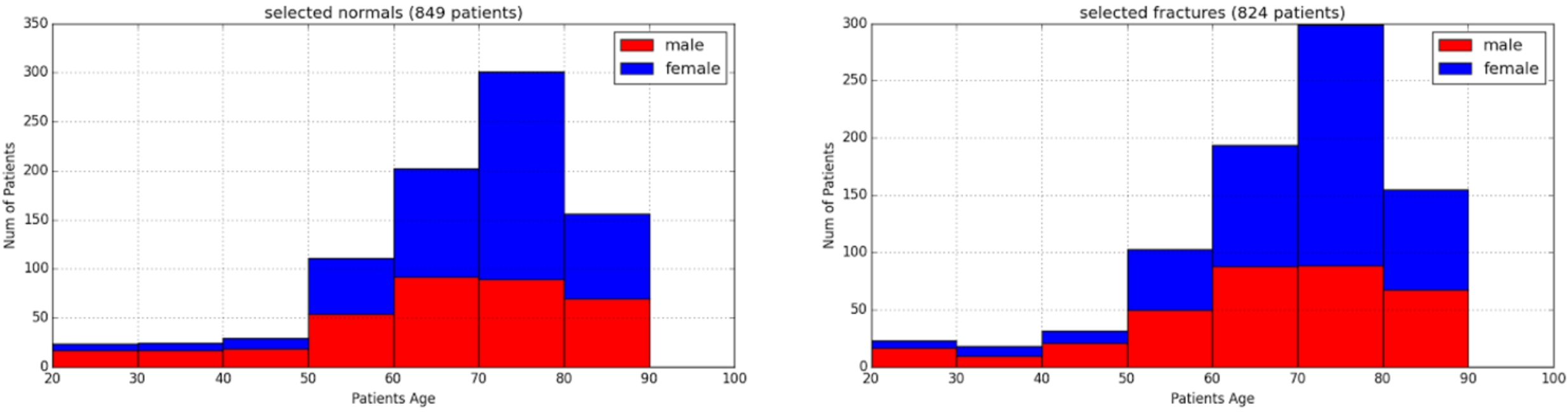}
   \end{tabular}
   \end{center}
   \caption[dataset] 
   { \label{fig:extracted_dataset} 
Extracted age and sex balanced data set: negative class (left) and positive class (right).  }
   \end{figure}

Figure \ref{fig:init_dataset} and Figure  \ref{fig:extracted_dataset} illustrate the initial and the extracted datasets properties respectively.

\section{Method}

To automatically detect VCF's from a provided CT Chest or Abdomen, we constructed an algorithmic pipeline of three stages: first, we segment the portion of the spine included in the provided CT and extract sagittal patches. We then classify the patches using a CNN and employ a RNN on the resulting vector of probabilities. The training phase was performed on the patches extracted using the algorithm below.

\subsection{Spine Segmentation and Patch Extraction}
The spine segmentation algorithmic goal is to enable deterministic sagittal section patches extracted along the vertebral column. These patches are used as input samples for training and inference. The use of the patches will be further explained in subsection \ref{cnn_classification}.

Several spine segmentation approaches have been described.  Ghosh et al [\citenum{ghosh2011automatic}] achieve spinal segmentation based upon the mid vertebral body in sagittal projection - a technique which has limited precision when applied to individuals with spinal scoliosis. To overcome limitations of variant spinal curvatures, we included a pre-processing step in generating a "virtual sagittal section" which first identifies and aligns the spinal cord in a straight cranial-caudal projection and then adjusts vertebral body location accordingly. We preferred this more relaxed segmentation approach for the task of VCF detection relative to several more accurate segmentation techniques for vertebrae [\citenum{yao2012detection},\citenum{kelm2013spine}] or discs [\citenum{ghosh2011automatic}, \citenum{yao2006automated}].

\begin{figure} [t!]
   \begin{center}
   \begin{tabular}{c} 
   \includegraphics[height=4cm]{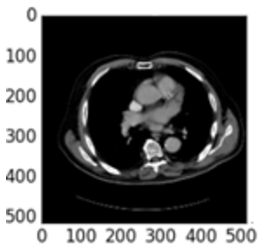}
   \includegraphics[height=4cm]{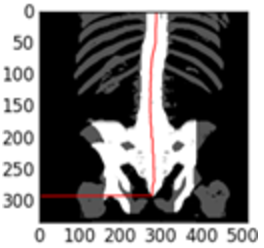}
   \includegraphics[height=4cm]{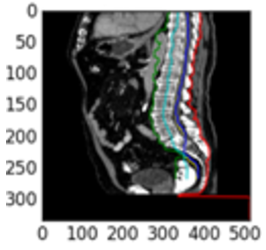}
   \includegraphics[height=4cm]{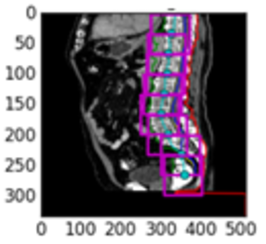}

   \end{tabular}
   \end{center}
   \caption[seg] 
   { \label{fig:seg} 
   The above set of images describes the flow of segmentation and patch extraction. (a) An Axial reconstruction CT Chest DICOM image obtained at the level of the heart with a thoracic vertebra in the lower (posterior) middle location on the image. (b) A coronal reconstruction of the axial skeleton and ribs created by a binarization manifold following the scanned individual's back, where HU values measure bones intensity.  The red line is the calculated position of the spinal cord. (c) A virtual sagittal section created by following the spinal cord. On this 2D section the vertebral column is segmented and its average width is measured. (d) Patches extracted along the vertebral column.

}
   \end{figure}

The “virtual sagittal section” creation is based on localization of the spinal cord in axial view, followed by tracking the back (posterior) of the individual along the spinal cord in coronal projection and bone intensity Hounsfield Unit (HU) threshold. The “virtual sagittal section” is used for segmentation of the vertebral column and extraction of sagittal patches of vertebrae to be learned and classified. The process of the algorithm is described in Figure \ref{fig:seg}.

This method was employed on the dataset described in section \ref{data} and was used to extract both positive and negative samples.  

   \begin{figure} [t!]
   \begin{center}
   \begin{tabular}{c} 
   \includegraphics[height=4.5cm]{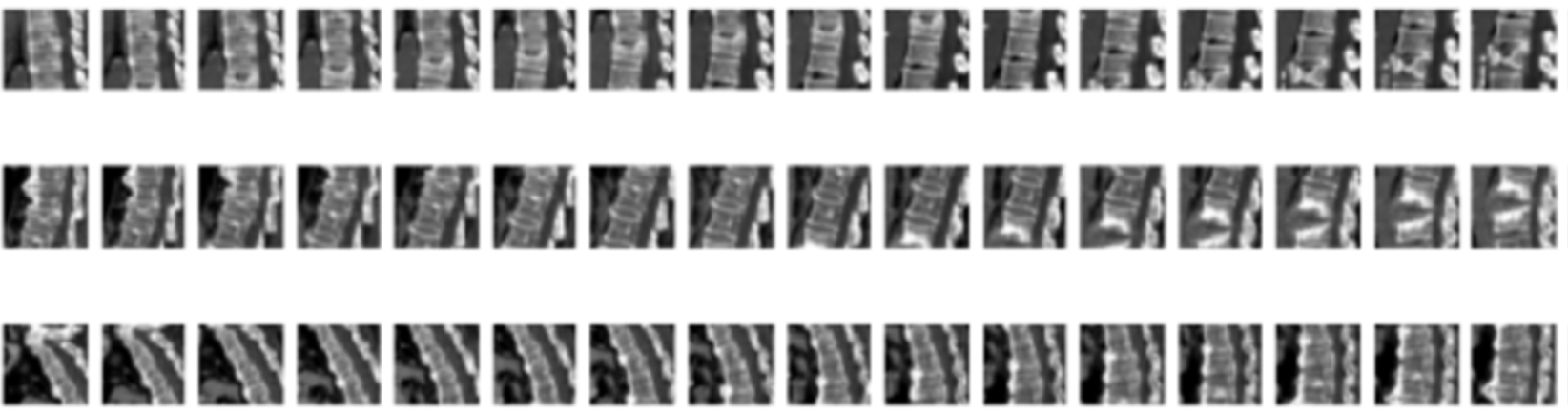}
   \end{tabular}
   \end{center}
   \caption[seg] 
   { \label{fig:seg_patches} 
This is the output of the segmentation algorithm. Each line corresponds to one CT scan. Note that each line is only a subset of the sequence of patches. }
   \end{figure} 

\subsection{Patch-based CNN}
We applied the present algorithm to the CT data set to extract patches along the vertebral column. Figure ~ֿֿֿ\ref{fig:vertebras} demonstrates the output of the segmentation algorithm which is an input to the CNN training phase. The patches were re-scaled to 32x32. We found this size ideal -  there were no significant improvements in our experiments for larger patch sizes such as 64x64 and 128x128. 15\% of all CT studies were sequestered as an exclusive validation set -  patches derived from any individual CT study reside in either the training or validation set but not both. The validation set held a balanced number of samples between the two classes. 

   \begin{figure} [t!]
   \begin{center}
   \begin{tabular}{c}
   \includegraphics[height=4cm]{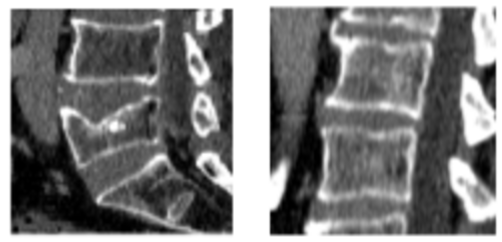}
   \end{tabular}
   \end{center}
   \caption[seg] 
   { \label{fig:vertebras} 
Patch containing a VCF (left). Right: Normal patch.}
   \end{figure} 

In contrast to other computer vision tasks where various data augmentation techniques are used in the training process, we opted not to apply any augmentation tools which might misrepresent the anatomic structure of the vertebral column. Instead applied only light rotations of -18 to 18 degrees, as the resulted transformation can naturally match the curvature and variability of the spine. 

\begin{figure} [t!]
   \begin{center}
   \begin{tabular}{c} 
   \includegraphics[height=5cm]{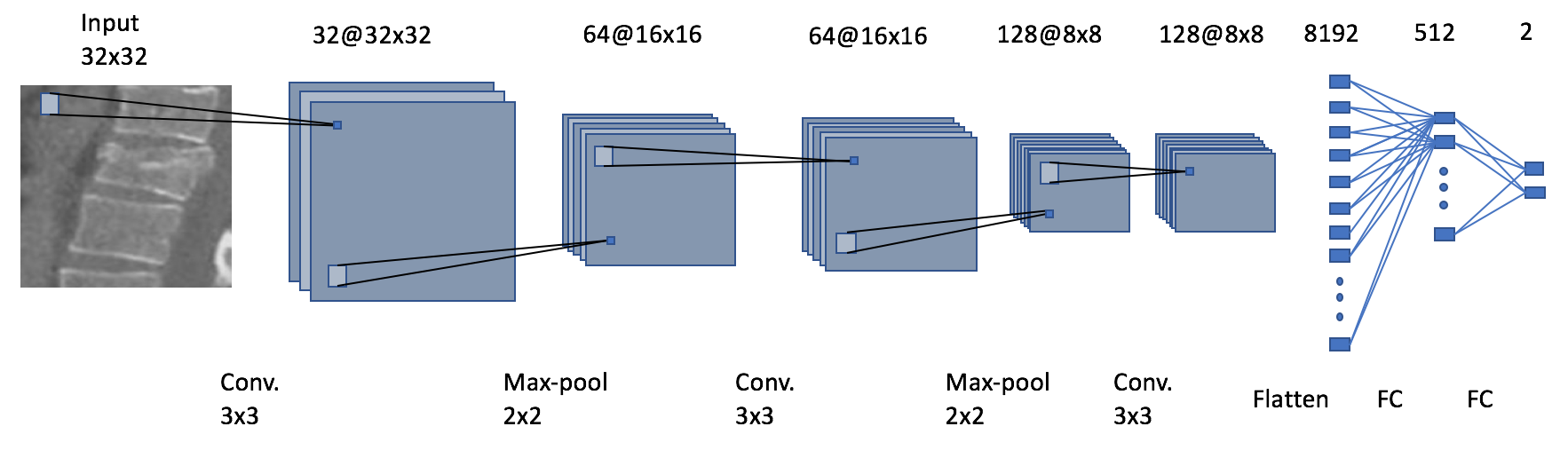}
   \end{tabular}
   \end{center}
   \caption[seg] 
   { \label{fig:cnn} 
CNN architecture used on input patches. The network was trained to predict whether a given patch contains a VCF.}
   \end{figure}

The CNN architecture employed is a variant based on VGG [\citenum{simonyan2014very}] adapted to 32x32 input. More specifically the network consists of the following layers: 
 32 3x3 convolutions, ReLU, 3x3 Max Pooling, 64 3x3 convolutions, ReLU, 64 3x3 convolutions, ReLU, 3x3 Max Pooling, 128 3x3 convolutions, ReLU, 128 3x3 convolutions, ReLU, 3x3 Max Pooling, a fully-connected layer of 512 dimensions, ReLU, Drop out, a fully-connected 2D layer, softmax. Figure \ref{fig:cnn} graphically demonstrates this architecture.

\subsection{Patch sequence classification} \label{cnn_classification}

RNN's have been a dominant tool in handling sequences for many tasks in fields such as video classification [\citenum{DBLP:journals/corr/NgHVVMT15}], language modeling [\citenum{mikolov2010recurrent}]and speech recognition [\citenum{graves2013speech}]. A popular approach is to use LSTM based RNN's [\citenum{hochreiter1997long}] which are more robust to long sequences as they don't suffer the vanishing and exploding gradient during training, which is a disadvantage for normal RNN's. 

\begin{figure} [t!]
   \begin{center}
   \begin{tabular}{c} 
   \includegraphics[height=6cm]{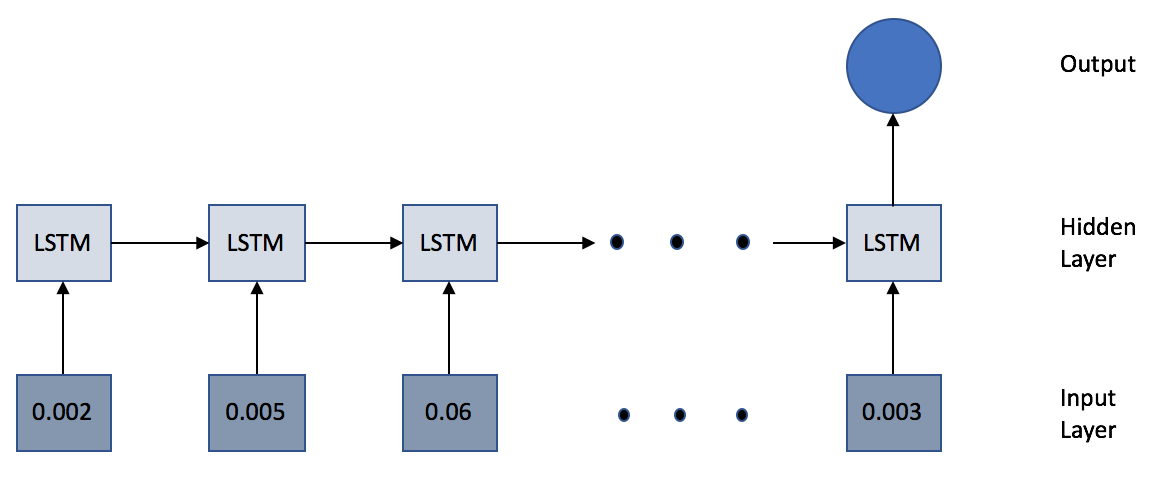}
   \end{tabular}
   \end{center}
   \caption[seg] 
   { \label{fig:rnn} 
RNN model used. The input is a CT scan represented by a vector of probabilities for VCF. The output is a prediction for whether the CT scan contains a VCF.}
   \end{figure}

Given a CT scan, we extract the sagittal vertebral patches and apply the CNN described to obtain a vector of probabilities. Figure \ref{fig:vertebras} illustrates a subset of a resulting vector of probabilities which represents a CT scan of a patient. The input vector size is not fixed; this is due to variance in physical attributes such as the height and the curvature of the spine. In addition, the vector of probabilities obtained is not a regular vector of features, but rather represents a sequence of predictions for fractures along the spine, therefore justifying an RNN based approach. Specifically, we employed a single layered LSTM with 128 cells, which were connected to a 2D output trained via the cross entropy loss. Figure \ref{fig:rnn} illustrates the RNN model used.
   \begin{figure} [t!]
   \begin{center}
   \begin{tabular}{c} 
   \includegraphics[height=2.5cm]{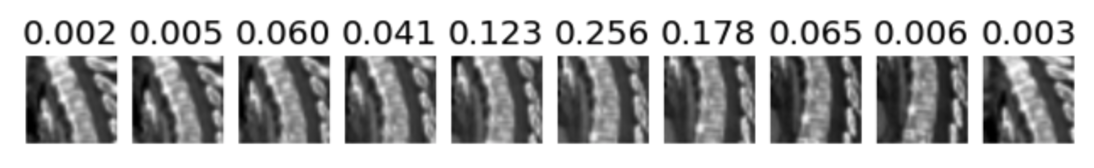}
   \end{tabular}
   \end{center}
   \caption[seg] 
   { \label{fig:vertebras} 
Employing the trained CNN on a given sequence of patches results in a vector of probabilities. Each probability represents the network prediction for VCF.  }
   \end{figure} 

\section{Results}
\begin{figure} [t!]
   \begin{center}
   \begin{tabular}{c} %% tabular useful for creating an array of images 
   \includegraphics[height=5cm]{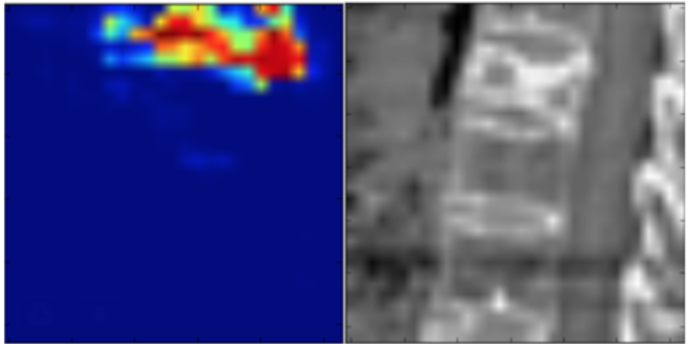}
   \end{tabular}
   \end{center}
   \caption[viz] 
%>>>> use \label inside caption to get Fig. number with \ref{}
   { \label{fig:visualization} 
Heat map visualization produced using the CNN trained classifier. The CNN focus is on the VCF.}
   \end{figure}

The CNN was trained for 15 epochs and reached 92.9\% accuracy over the validation set. Visualizing training samples such in Figure ~\ref{fig:visualization}, demonstrates that the trained CNN learned to focus on VCF's. We then applied the trained CNN to the entire training and validation data sets to generate vectors for the RNN based classifier. The RNN was trained on the training data set and evaluated upon the validation vectors, resulting in 89.1\% accuracy, 83.9\% sensitivity and 93.8 specificity.
\section{Conclusions}
VCF's are clinically important findings which are readily identifiable but routinely overlooked by radiologists. Only 13\%-16\% of retrospectively confirmed VCF’s are actually reported at the time of CT study interpretation. This is likely due to the fact that they are considered incidental findings relative to the primary reason for which a given CT study was performed. In this work we present a robust algorithm to detect VCF's in CT scans of the Chest and/or Abdomen. First, a deterministic segmentation algorithm extracts patches along the patient’s vertebral column. Then these patches are classified using a CNN followed by RNN classifier on the resulting vector of probabilities. To the best of our knowledge this work is the first to address the problem of VCF detection using deep learning methodologies. This application may serve to increase the diagnostic sensitivity for VCF's in routinely acquired CT's, potentially triggering preventative measures to reduce the rate of future hip fracture. 

\acknowledgments % equivalent to \section*{ACKNOWLEDGMENTS}       
This research was supported by Zebra Medical Vision.

% References
\bibliography{report} % bibliography data in report.bib
\bibliographystyle{spiebib} % makes bibtex use spiebib.bst

\end{document}